\documentclass{article}
\usepackage{iclr2027_conference,times}

\usepackage{amsmath,amssymb,booktabs,array,graphicx}
\usepackage{colortbl}
\usepackage{algorithm}
\usepackage{algorithmic}
\usepackage{hyperref}
\usepackage{url}
\usepackage{here}
\usepackage{tcolorbox}

\definecolor{vrrlink}{RGB}{20,76,145}
\definecolor{vrrshade}{gray}{0.955}
\definecolor{vrrcolumn}{RGB}{229,238,250}
\hypersetup{
  colorlinks=true,
  linkcolor=vrrlink,
  citecolor=vrrlink,
  urlcolor=vrrlink
}

\newtcolorbox{keyresult}{
  colback=vrrshade,
  colframe=vrrshade,
  boxrule=0pt,
  arc=0pt,
  left=6pt,
  right=6pt,
  top=5pt,
  bottom=5pt,
  before skip=7pt,
  after skip=7pt
}

\title{LLM-as-an-Improver: Turning Verification into Better Candidates}

\author{Akiyoshi Tomihari 
\\
Fujitsu Limited \\  
Department of Computer Science\\
The University of Tokyo\\
\And 
Yuma Ichikawa 
\\
Fujitsu Limited \\
RIKEN center for AIP
}

\newcommand{\selector}{\textsc{Verify-Select}}
\newcommand{\lav}{\textsc{LAV}}
\newcommand{\sat}{\textsc{Sat}}
\newcommand{\viol}{\textsc{Viol}}
\newcommand{\unk}{\textsc{Unknown}}

\iclrpreprintcopy 

\begin{document}

\maketitle

\begin{abstract}
Verifier-based selection improves LLM performance by generating multiple candidate solutions and using a verifier to select the most promising one. However, existing methods typically treat verification only as a ranking step and discard its feedback once a fixed candidate pool has been evaluated. In this paper, we ask whether verification can also improve the candidate set itself. To this end, we introduce LLM-as-an-Improver and propose Verify--Repair--Reselect (VRR), which uses verification feedback to generate and reselect improved candidates. VRR retains the initial winner while conditionally generating three complementary alternatives: repaired versions of the winner and runner-up, and a solution based on a new approach. It filters invalid and duplicate candidates using only inference-time information and then reselects the final answer under the original evaluation criteria. Across diverse models and code-generation and reasoning benchmarks, VRR improves over fixed-pool verifier-based selection in many settings and can recover correct solutions even when all candidates in the initial pool are incorrect. These results highlight a broader role for LLMs as improvers: verification feedback can not only select among existing solutions but also construct stronger candidates beyond the initial pool.
\end{abstract}


\section{Introduction}

Test-time scaling improves the capabilities of large language models (LLMs) by allocating additional inference compute to candidate generation and selection.
Generation methods increase solution coverage through repeated sampling or diverse reasoning paths~\citep{wang2023selfconsistency,brown2024largelanguagemonkeys}, although the effectiveness of additional compute depends on both problem difficulty and its allocation~\citep{snell2025scaling}.
Some methods use verifiers, which attempt to identify the strongest candidate using outcome, process, or LLM-based judgments~\citep{cobbe2021training,lightman2023verify,kwok2026llmverifier}.
In these methods, generation determines which candidates are available, and verification is used only to rank them.

However, these methods do not fully exploit the potential of verification.
A verifier may identify violated requirements, unsupported reasoning steps, failing cases, and unresolved claims, yet selection-only inference compresses this information into a ranking and then discards it.
Moreover, ranking cannot repair the search space: if every candidate in the initial pool is incorrect, even an oracle selector must fail.
This leads to the central research question of this paper: \emph{Can verification also guide the construction of better candidates?}

We refer to this expanded role as LLM-as-an-Improver: rather than using verification only to choose among existing candidates, the model reuses diagnostic evidence to construct additional candidates.
We propose Verify--Repair--Reselect (VRR), a test-time method that uses structured verification results both to rank candidates and to guide additional generation.
VRR first selects and retains the top-ranked initial candidate.
When the initial pool appears unreliable, it generates three alternatives: a local repair of the winner, a repair of the runner-up, and a solution using a new approach.
It removes invalid and duplicate alternatives, then selects the final answer from the surviving alternatives and the initial winner using the same evaluation criteria.
Unlike single-trajectory refinement, VRR treats verifier feedback as a signal for expanding the choice set: it preserves the initial winner, creates complementary alternatives, and subjects all surviving candidates to reselection rather than accepting a revision automatically.


Across two models and eight benchmarks, VRR improves over fixed-pool \lav{} selection in nine of 16 settings and, crucially, recovers correct solutions on LiveCodeBench problems for which every initial candidate is incorrect.

\begin{figure*}[t]
  \centering
  \includegraphics[width=\textwidth]{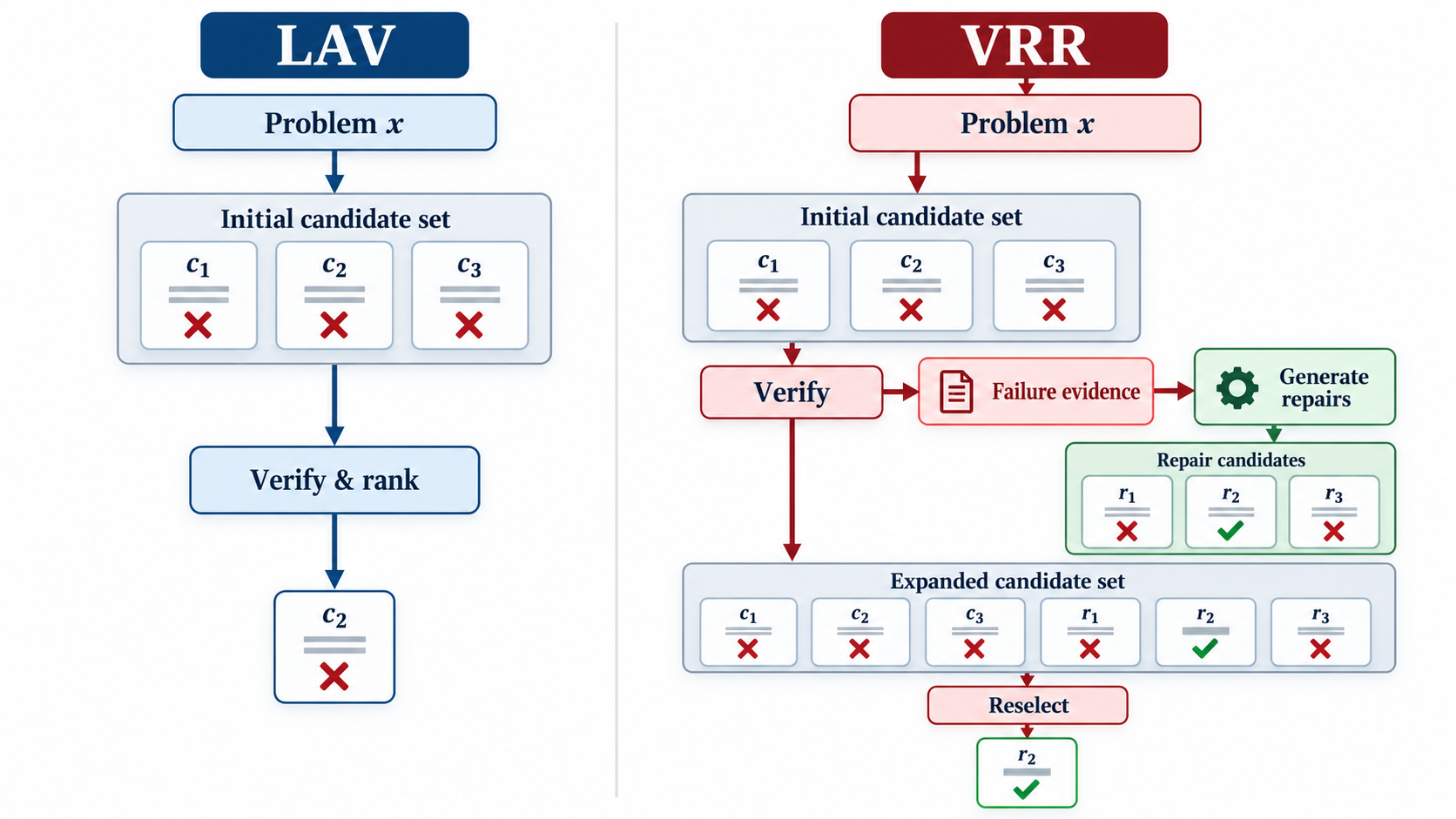}
  \caption{Two uses of verification at test time. LAV uses verification to rank a fixed initial set. VRR additionally turns diagnostic verification evidence into new candidates before reselection, making recovery possible even when every initial candidate is incorrect.}
  \label{fig:lav-vrr-overview}
\end{figure*}

Figure~\ref{fig:lav-vrr-overview} illustrates the central distinction: selection-only inference uses verification to change the ranking, whereas evidence-guided expansion also uses it to change the candidates being ranked.

\section{Problem Setting}
\label{sec:problem}

\subsection{Verification for Candidate Ranking}

Candidate generation and verifier-based selection play distinct roles in test-time inference: generation determines which solutions are available, while verification determines which candidate is returned.
Pass@$k$ measures whether at least one of $k$ candidates is correct~\citep{brown2024largelanguagemonkeys}; verifier-based selection attempts to convert this candidate coverage into final-answer accuracy.

Formally, let $x$ denote a task and let $\mathcal{C}_{\mathrm{init}}=\{c_1,\ldots,c_N\}$ be the initial candidate set produced by an LLM generator.
Each $c_i$ is a candidate solution containing a final answer and, when appropriate, its reasoning process or code.
In selection-only methods, a verifier $V$ scores or compares these candidates and returns one of them:
\begin{equation}
  \hat{c}_{\mathrm{select}}
  =\operatorname{Select}_{V}
  \left(x,\mathcal{C}_{\mathrm{init}}\right),
  \qquad
  \hat{c}_{\mathrm{select}}\in\mathcal{C}_{\mathrm{init}}.
  \label{eq:selection-only}
\end{equation}
This formulation includes outcome verifiers, process verifiers, generative verifiers, and LLM-based judges~\citep{cobbe2021training,lightman2023verify,zhang2025generative,kwok2026llmverifier}.
These assessments may be rich, but selection-only methods use them for a single purpose: ranking an already generated set.

\subsection{Why Ranking Alone Is Insufficient}

Let $Y_x(c)\in\{0,1\}$ denote correctness under the task's evaluation rule.
This label is used only for evaluation and is unavailable to the inference procedure.
For task $x$, the initial-pool coverage indicator is
\begin{equation}
  H_0(x)=\max_{c\in\mathcal{C}_{\mathrm{init}}}Y_x(c).
  \label{eq:selection-ceiling}
\end{equation}
We define the initial-pool oracle accuracy as
\begin{equation}
  A_{\mathrm{oracle}} := \mathbb{E}[H_0(x)],
  \label{eq:oracle-accuracy}
\end{equation}
where expectations are taken over the task distribution and any randomness in candidate generation and inference.
This is the accuracy of an oracle selector that returns a correct candidate whenever the initial pool contains one.
Because $Y_x(\hat c_{\mathrm{select}})\leq H_0(x)$ for every task $x$, any selection-only method satisfies
\begin{equation}
  \mathbb{E}[Y_x(\hat c_{\mathrm{select}})]
  \leq A_{\mathrm{oracle}}.
  \label{eq:oracle-ceiling}
\end{equation}
For a fixed pool, $H_0(x)$ is a post-evaluation coverage indicator, not an estimate of selected-candidate correctness, and is unavailable at inference time.
This ceiling requires no independence assumption.
It separates candidate \emph{coverage} from the selector's ability to identify a correct candidate.
The limitation therefore lies not in verifier strength but in using verification only to rearrange candidates.
A stronger verifier can approach the ceiling, but only candidate generation can raise it.
Diagnostic content produced during verification offers a task-specific signal for doing so.

\subsection{From Ranking to Candidate Improvement}
\label{sec:general-expansion}

We therefore extend the role of verification rather than replace verifier-based selection.
Formally, verification produces both a ranking signal $r_V$ and diagnostic evidence $D_V$:
\begin{equation}
  (r_V,D_V)
  =\operatorname{Verify}_{V}(x,\mathcal{C}_{\mathrm{init}}).
  \label{eq:diagnostics}
\end{equation}
Here, $\operatorname{Verify}_{V}$ evaluates the initial candidates for task $x$ and returns both the information used to rank them, $r_V$, and candidate-level diagnostic evidence, $D_V$.
Selection-only inference consumes $r_V$ and leaves $D_V$ unused.
The diagnostic evidence is also fed back into generation, including criterion-level critiques, counterexamples, and unresolved claims:
\begin{equation}
  \mathcal{R}=G(x,\mathcal{C}_{\mathrm{init}},D_V),
  \qquad
  \mathcal{R}_{\mathrm{pass}}
  =\operatorname{Screen}(x,\mathcal{R};\mathcal{C}_{\mathrm{init}}),
  \label{eq:general-generation}
\end{equation}
where $G$ denotes a candidate generator that conditions on the task, the initial candidates, and the verification evidence to produce a set $\mathcal{R}$ of additional candidates.
$\operatorname{Screen}$ applies inference-time validity checks and removes candidates that duplicate an initial or already retained candidate, yielding the admissible set $\mathcal{R}_{\mathrm{pass}}$.
Screening is pool-dependent because duplicate status depends on the other candidates.
Passing the screen establishes eligibility for reselection, not correctness.
A retention policy then chooses a subset
$\mathcal{K}=K(\mathcal{C}_{\mathrm{init}},D_V)$ of initial candidates.
The final output is
\begin{equation}
  \hat c_{\mathrm{expand}}
  =\operatorname{Select}_{V}
  \left(x,\mathcal{K}\cup\mathcal{R}_{\mathrm{pass}}\right).
  \label{eq:general-expansion}
\end{equation}
The expanded set is then verified again, so the same capability supports both candidate improvement and final ranking.
Selection-only inference corresponds to the special case that ignores $D_V$, generates no new candidates, and sets $\mathcal{K}=\mathcal{C}_{\mathrm{init}}$.
Candidate expansion makes recovery possible even when $H_0(x)=0$.


\section{Verify--Repair--Reselect}
\label{sec:method}

VRR first ranks the initial candidates, reuses the same criterion-level evidence to guide candidate improvement, and then performs final reselection.
It consists of five stages:
\begin{enumerate}
  \item Derive evaluation criteria from the problem statement.
  \item Evaluate the initial candidates and select the top-ranked candidate.
  \item Use the evaluation results to decide whether to trigger repair.
  \item If triggered, generate three candidates with complementary roles.
  \item Screen the new candidates and reselect under the original criteria.
\end{enumerate}

\subsection{Creating Evaluation Criteria}
\label{sec:criteria}

For each task, we derive a fixed finite collection of criteria from the problem statement $x$:
\begin{equation}
  \Gamma(x)
  =\mathcal{O}_{\mathrm{hard}}(x)
  \cup\mathcal{O}_{\mathrm{soft}}(x),~~~|\mathcal{O}_{\mathrm{hard}}(x)| \ge 1.
  \label{eq:criteria-portfolio}
\end{equation}
Each evaluation item $o_j$ specifies what to judge, what evidence may be used, possible counterexamples, and a score initialization parameter $\pi_j$.

Hard items are shared within a dataset and treated as necessary conditions for correctness.
For example, code tasks consider conformance to the specification, functional correctness, syntax errors, failures on public tests, and complexity violations.
By contrast, soft items are generated by the LLM separately for each problem based on its problem statement and are used only to break near-ties in the hard-item scores.

\subsection{Checking Candidates by Evaluation Item}
\label{sec:evaluation}

Before evaluating candidate $c_i$, we organize the information available to the verifier as
\begin{equation}
  E_i\leftarrow\operatorname{BuildEvidenceGraph}(x,c_i).
  \label{eq:build-evidence-graph}
\end{equation}
Here, $E_i$ collects the information used to evaluate candidate $c_i$.
It organizes the problem statement $x$, candidate $c_i$, and public information available at inference time into a form that the verifier can reference.

For general reasoning tasks, the candidate response is divided into reasoning steps and a final answer, and each element is linked to its location in the response.
For code tasks, $E_i$ records the candidate code or patch together with syntax checks, compilation results, and the results of designated public tests.
For tasks involving file changes, it also records the changed locations and target files.
Each piece of information is assigned a source location or identifier so that the verifier can cite a specific part of the candidate or a public test result as the basis for its judgment.

For candidate $i$ and evaluation item $j$, the verifier's $t$-th judgment returns
\begin{equation}
  \boldsymbol{q}_{ij}^{(t)}
  =\left(q^{\sat,(t)}_{ij},q^{\viol,(t)}_{ij},q^{\unk,(t)}_{ij}\right)
  \in\Delta^2.
\end{equation}
Here, $\Delta^2$ is the probability simplex over the three labels.
The verifier also returns $g_{ij}^{(t)}\in\{0,1\}$, which indicates whether it grounds the judgment in the available evidence.
The initial batched evaluation corresponds to $t=1$, and adaptive evaluations produce up to $T_{ij}$ judgments for each entry.
The \unk{} state separates missing evidence from evidence that supports satisfaction.
We define the judgment weight as
\begin{equation}
  w_{ij}^{(t)}
  =g_{ij}^{(t)}\left(1-q^{\unk,(t)}_{ij}\right)
  \left(1-\frac{H(\boldsymbol{q}_{ij}^{(t)})}{\log 3}\right).
  \label{eq:judgment-weight}
\end{equation}
Here, $H$ denotes Shannon entropy.
This weight lies in $[0,1]$; it is zero for an ungrounded, fully unresolved, or uniform judgment and otherwise discounts reported uncertainty.

We convert the weighted judgments into a criterion-level verification score.
Let $\pi_j\in(0,1)$ be an initialization parameter and $\epsilon>0$ a numerical smoothing constant.
Define
\begin{equation}
  \ell_{ij}
  =\operatorname{logit}(\pi_j)
  +\sum_{t=1}^{T_{ij}} w^{(t)}_{ij}
  \log\frac{q^{\sat,(t)}_{ij}+\epsilon}
            {q^{\viol,(t)}_{ij}+\epsilon}.
  \label{eq:verification-log-odds}
\end{equation}
We map this value to $(0,1)$ as
\begin{equation}
  p_{ij}=\operatorname{sigmoid}(\ell_{ij}).
  \label{eq:criterion-score}
\end{equation}
Equations~\ref{eq:verification-log-odds}--\ref{eq:criterion-score} define a bounded, confidence-weighted score for ranking and triggering.
The verifier's reported ratio $q^{\sat}/q^{\viol}$ is not an observed likelihood ratio, repeated judgments may be dependent, and evaluation criteria may overlap.
The log-odds form is therefore an aggregation heuristic, not a probabilistic guarantee.

We aggregate the hard items as
\begin{equation}
  S_i=\prod_{o_j\in\mathcal{O}_{\mathrm{hard}}}p_{ij}.
  \label{eq:success-score}
\end{equation}
The product is a multiplicative aggregation rule that penalizes a low score on any required item.
It is not an established probability that all hard criteria hold, and its scale depends on the number and definition of the criteria.
Soft items break ties only when two hard-item scores differ by at most $\tau_{\mathrm{soft}}$.

\subsection{Adaptive Verification and Initial Selection}
\label{sec:initial-selection}

The verifier first judges every candidate--item entry once in a batch.
It then directs the remaining budget to entries that can change the final ranking.
We use the following acquisition score:
\begin{equation}
  A_{ij}=I_{ij}\,H_2(p_{ij})\,D_i/\kappa_{ij}.
  \label{eq:acquisition}
\end{equation}
Here, $I_{ij}$ indicates whether item $j$ can change the all-hard-items decision for candidate $i$.
The Bernoulli entropy $H_2(p_{ij})$ measures uncertainty.
The factor $D_i$ prioritizes candidates near the selection boundary.
The term $\kappa_{ij}$ denotes the expected judgment cost.
At each step, the verifier judges the entry with the largest $A_{ij}$.

The selector then compares the top two candidates directly.
It evaluates both presentation orders, $A=c_a,B=c_b$ and $A=c_b,B=c_a$, for every evaluation item \citep{wang2024unfairevaluators}.
Hard-item comparison evidence updates the candidate scores.
Soft items affect the decision only when the hard scores are close.
This comparison selects the initial winner $c_{i^*}$.
Let $\boldsymbol{\rho}$ denote the resulting ranking, where $c_{\rho_1}=c_{i^*}$ and $c_{\rho_2}$ is the runner-up.

\begin{algorithm}[t]
\caption{Verify--Repair--Reselect (VRR)}
\label{alg:vrr}
\begin{algorithmic}[1]
\REQUIRE Task $x$; initial candidates $\mathcal{C}_{\mathrm{init}}$; verifier $V$; repair generator $G$; budgets $B_{\mathrm{init}},B_{\mathrm{reselect}}$; thresholds $\tau_{\mathrm{sat}},\tau_{\mathrm{viol}},\tau_{\mathrm{score}},\tau_{\mathrm{unk}},K_{\mathrm{unk}}$
\ENSURE Final candidate $\hat{c}$
\STATE $\Gamma\leftarrow\operatorname{CreateCriteria}(x)$
\FORALL{$c_i\in\mathcal{C}_{\mathrm{init}}$}
  \STATE $E_i\leftarrow\operatorname{BuildEvidenceGraph}(x,c_i)$
\ENDFOR
\STATE $(c_{i^*},\boldsymbol{S},L,\boldsymbol{\rho})
       \leftarrow\selector(x,\mathcal{C}_{\mathrm{init}},\{E_i\},\Gamma,V,
       B_{\mathrm{init}})$
\STATE $u\leftarrow
       \#\{(i,j):o_j\in\mathcal{O}_{\mathrm{hard}},
       q^{\unk,(1)}_{ij}\geq\tau_{\mathrm{unk}}\}$
\IF{$\max_i S_i\geq\tau_{\mathrm{score}}$ \AND $u<K_{\mathrm{unk}}$}
  \STATE \textbf{return} $c_{i^*}$
\ENDIF
\STATE $\mathcal{R}\leftarrow
       \operatorname{Repair}(x,c_{\rho_1},c_{\rho_2},L,G)$
\COMMENT{selected, runner-up, and new-approach}
\STATE $\mathcal{R}_{\mathrm{pass}}
       \leftarrow\{c\in\mathcal{R}:\operatorname{Screen}(x,c)=1\}$
\IF{$\mathcal{R}_{\mathrm{pass}}=\varnothing$}
  \STATE \textbf{return} $c_{i^*}$
\ENDIF
\STATE $\mathcal{C}_{\mathrm{reselect}}\leftarrow
       \{c_{i^*}\}\cup\mathcal{R}_{\mathrm{pass}}$
\FORALL{$c\in\mathcal{R}_{\mathrm{pass}}$}
  \STATE $E_c\leftarrow\operatorname{BuildEvidenceGraph}(x,c)$
\ENDFOR
\STATE $\mathcal{E}_{\mathrm{reselect}}\leftarrow
       \{E_{i^*}\}\cup\{E_c:c\in\mathcal{R}_{\mathrm{pass}}\}$
\STATE $(\hat{c},\cdot,\cdot,\cdot)
       \leftarrow\selector(x,\mathcal{C}_{\mathrm{reselect}},
       \mathcal{E}_{\mathrm{reselect}},\Gamma,V,B_{\mathrm{reselect}})$
\STATE \textbf{return} $\hat{c}$
\end{algorithmic}
\end{algorithm}

\subsection{Summarizing Results and Deciding Whether to Generate Repairs}
\label{sec:generation-condition}

We summarize the initial evaluation in $L$, a diagnostic record that makes information produced for ranking reusable by the generator.
For each candidate, $L$ records the evaluation items, aggregate score, uncertainty interval, and initial rank.
Using the initial batched judgments, $L$ records strengths with $q^{\sat,(1)}_{ij}\geq\tau_{\mathrm{sat}}$, potential issues with $q^{\viol,(1)}_{ij}\geq\tau_{\mathrm{viol}}$, and unresolved items with $q^{\unk,(1)}_{ij}\geq\tau_{\mathrm{unk}}$.
Potential issues are not treated as confirmed errors; they are passed to the repair generator as items that require re-examination.

We decide whether to generate additional candidates using two conditions:
\begin{align}
  \mathrm{low\mbox{-}pool\mbox{-}score}
  &\iff\max_i S_i<\tau_{\mathrm{score}},\\
  \mathrm{high\mbox{-}unknown\mbox{-}count}
  &\iff
  \#\left\{(i,j):o_j\in\mathcal{O}_{\mathrm{hard}},
  q^{\unk,(1)}_{ij}\geq\tau_{\mathrm{unk}}\right\}
  \geq K_{\mathrm{unk}}.
\end{align}
The thresholds $\tau_{\mathrm{sat}}$ and $\tau_{\mathrm{viol}}$ determine which strengths and potential issues enter $L$.
The threshold $\tau_{\mathrm{score}}$ defines a low candidate score.
The threshold $\tau_{\mathrm{unk}}$ defines an unresolved judgment.
The integer $K_{\mathrm{unk}}$ defines the required number of unresolved judgments.
Appendix~\ref{app:experimental-details} specifies all thresholds.
The repair trigger is
\begin{equation}
  \mathcal{T}=\mathrm{low\mbox{-}pool\mbox{-}score}
  \lor\mathrm{high\mbox{-}unknown\mbox{-}count}.
\end{equation}
The first condition holds when every candidate score is below $\tau_{\mathrm{score}}$.
The second holds when at least $K_{\mathrm{unk}}$ hard-item judgments from the initial batch remain unresolved.
If neither condition holds, VRR returns $c_{i^*}$ without additional generation.

\subsection{Three Types of Additional Generation}
\label{sec:additional-generation}

When $\mathcal{T}$ is true, VRR changes the role of the initial verification result from a ranking signal into a generation signal and produces three additional candidates.
The three roles cover both local errors and failures of the underlying approach, although they do not guarantee semantically distinct outputs.

\noindent\textbf{Selected-candidate repair.}
The generator re-examines potential issues in the winner while preserving the remaining parts where possible.

\textbf{Runner-up repair.}
It separately repairs the second-ranked candidate to retain a potentially different starting point.

\textbf{New-approach solution.}
It identifies possible failure conditions from the winner and its evaluation, then solves the task from scratch rather than modifying that candidate.

Each prompt contains the problem, its source candidate, and the corresponding evaluation summary, but no correct or reference answer, private-test result, or correctness label.
Because reported issues are verifier estimates rather than confirmed errors, the generator re-examines them and ignores unsupported feedback.
Together, the roles favor minimal repair, an alternative starting point, and a distinct approach instead of repeatedly refining one trajectory \citep{madaan2023selfrefine,shinn2023reflexion}.

\subsection{Screening Additional Candidates}
\label{sec:screening}

VRR screens each new candidate before reselection.
For code tasks, we retain a candidate only if we can extract a complete artifact, the artifact compiles, and it passes every designated public test.
We also require its normalized code hash to differ from those of all initial and retained candidates.
For multiple-choice and free-form reasoning, we require a nonempty and nontruncated response with inspectable reasoning and an extractable final answer in the specified format.
We reject normalized duplicates.

These checks do not determine whether a candidate is correct.
A program that passes public tests can still fail private tests, and a well-formed reasoning response can still be logically incorrect.
The reselection step determines whether an additional candidate is better than the initially selected candidate.

\subsection{Reselection Under the Same Criteria}
\label{sec:reselection}

If at least one new candidate passes screening, VRR forms
\begin{equation}
  \mathcal{C}_{\mathrm{reselect}}
  =\{c_{i^*}\}\cup\mathcal{R}_{\mathrm{pass}}
  \subseteq\mathcal{C}_{\mathrm{init}}\cup\mathcal{R}_{\mathrm{pass}},
  \qquad
  2\leq|\mathcal{C}_{\mathrm{reselect}}|\leq 4.
  \label{eq:reselection-pool}
\end{equation}
We exclude the other initial candidates to reserve the evaluation budget for verification-guided alternatives.
We retain $c_{i^*}$ so that repair does not remove the initial winner from consideration.
Thus, candidate generation expands the search space, but the active reselection pool need not contain every initial candidate or be larger than the initial pool.
This set inclusion does not guarantee that reselection will choose the winner when it is correct.

We reuse the evaluation criteria $\Gamma(x)$ created before inspecting the initial candidates and change only the candidate set.
For every candidate in $\mathcal{C}_{\mathrm{reselect}}$, we repeat batched judgments, adaptive additional judgments, and pairwise comparisons in both presentation orders.

Algorithm~\ref{alg:vrr} summarizes the complete inference procedure.

\section{Experiments}

We evaluate whether VRR improves aggregate accuracy over fixed-pool \lav{} selection and whether it can recover correct solutions when all candidates in the initial pool are incorrect.
We use Gemma 4 31B IT \citep{gemmateam2026gemma4} and Qwen3-32B \citep{qwenteam2025qwen3}; within each experiment, the same model performs generation, verification, repair, and reselection.
The code benchmarks are LiveCodeBench \citep{jain2024livecodebench}, HumanEval+, and MBPP \citep{liu2023evalplus}; the reasoning benchmarks are MMLU-Pro \citep{wang2024mmlupro}, GPQA Diamond \citep{rein2023gpqa}, AIME 2025 \citep{zhang2025aime25}, AMC 2023 \citep{mathai2025amc23}, and MATH-500 \citep{lightman2023verify}.
Initial pools contain five candidates for each code problem and three for each reasoning problem, and all methods receive the same frozen pool.
The Single baseline returns its first candidate, whereas \lav{} selects from the pool without additional generation.
Correctness labels, reference answers, and private tests are withheld until the final answer is fixed.
Detailed settings and evaluation rules are provided in Appendix~\ref{app:experimental-details},
and the generation, verification, repair, and \lav{} prompt templates are
provided in Appendix~\ref{app:prompts}.

\subsection{Accuracy Across Eight Benchmarks}
\label{sec:exp-overall}

\begin{table}[tbp]
  \centering
  \small
  \caption{Accuracy (\%) on frozen initial pools. Bold marks the highest accuracy, including ties; $\Delta$ is VRR minus \lav{} in percentage points.}
  \label{tab:overall-results}
  \resizebox{\textwidth}{!}{%
  \begin{tabular}{lrrrrrrrr}
    \toprule
    & \multicolumn{4}{c}{Gemma 4 31B IT}
    & \multicolumn{4}{c}{Qwen3-32B} \\
    \cmidrule(lr){2-5}\cmidrule(lr){6-9}
    Benchmark & Single & \lav{} & \cellcolor{vrrcolumn}VRR & $\Delta$
              & Single & \lav{} & \cellcolor{vrrcolumn}VRR & $\Delta$ \\
    \midrule
    LiveCodeBench & 83.22 & 87.58 & \cellcolor{vrrcolumn}\textbf{88.91} & +1.33
                  & 78.10 & 78.58 & \cellcolor{vrrcolumn}\textbf{79.53} & +0.95 \\
    HumanEval+    & 95.12 & \textbf{95.73} & \cellcolor{vrrcolumn}95.12 & -0.61
                  & 82.93 & 84.76 & \cellcolor{vrrcolumn}\textbf{85.37} & +0.61 \\
    MBPP          & 82.01 & 81.75 & \cellcolor{vrrcolumn}\textbf{83.07} & +1.32
                  & 75.40 & \textbf{76.46} & \cellcolor{vrrcolumn}76.19 & -0.27 \\
    MMLU-Pro      & 85.70 & 85.65 & \cellcolor{vrrcolumn}\textbf{85.75} & +0.10
                  & 75.05 & 75.80 & \cellcolor{vrrcolumn}\textbf{76.80} & +1.00 \\
    GPQA Diamond  & 78.79 & \textbf{80.30} & \cellcolor{vrrcolumn}\textbf{80.30} & 0.00
                  & 52.53 & 53.03 & \cellcolor{vrrcolumn}\textbf{57.07} & +4.04 \\
    AIME          & 73.33 & \textbf{76.67} & \cellcolor{vrrcolumn}\textbf{76.67} & 0.00
                  & \textbf{23.33} & \textbf{23.33} & \cellcolor{vrrcolumn}\textbf{23.33} & 0.00 \\
    AMC           & \textbf{97.50} & \textbf{97.50} & \cellcolor{vrrcolumn}\textbf{97.50} & 0.00
                  & 70.00 & 75.00 & \cellcolor{vrrcolumn}\textbf{77.50} & +2.50 \\
    MATH-500      & 98.60 & 98.40 & \cellcolor{vrrcolumn}\textbf{98.80} & +0.40
                  & 87.00 & \textbf{89.20} & \cellcolor{vrrcolumn}88.60 & -0.60 \\
    \bottomrule
  \end{tabular}
  }
\end{table}

Table~\ref{tab:overall-results} reports final-answer accuracy on the shared frozen pools.
With Gemma, VRR outperforms \lav{} on four benchmarks, ties on three, and underperforms on one; its largest gains are 1.33 points on LiveCodeBench and 1.32 on MBPP, while HumanEval+ decreases by 0.61.
With Qwen, VRR outperforms \lav{} on five benchmarks, ties it on one, and underperforms it on two; the largest gains are 4.04 points on GPQA Diamond and 2.50 on AMC, while its accuracy is lower by 0.27 points on MBPP and 0.60 points on MATH-500.
Overall, VRR outperforms \lav{} in nine of 16 settings, ties it in four, and underperforms it in three, motivating separate consideration of useful candidate generation and reliable reselection; these runs do not establish statistical significance or consistency across seeds.

\subsection{LiveCodeBench: Recovery, Coverage, and Missed Solutions}
\label{sec:exp-lcb}

\begin{figure}[tbph]
  \centering
  \includegraphics[width=\textwidth]{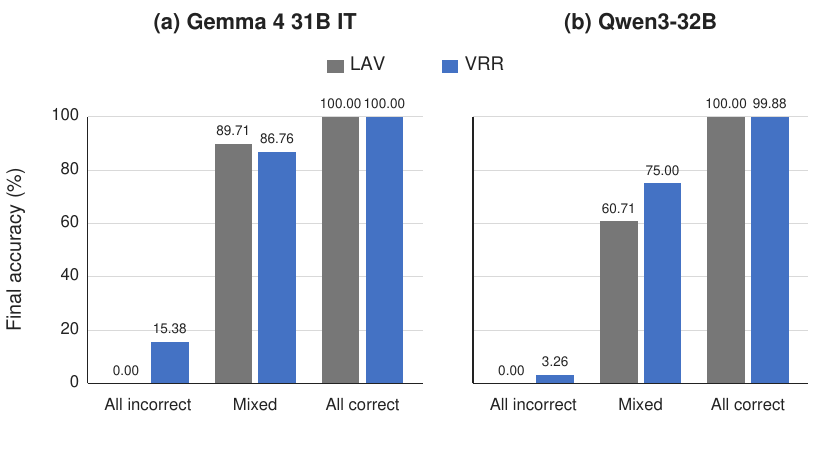}
  \caption{Final accuracy (\%) on LiveCodeBench, grouped by the
  composition of the five-candidate initial pool.}
  \label{fig:lcb-initial-accuracy-strata}
\end{figure}

In Figure~\ref{fig:lcb-initial-accuracy-strata}, we stratify all 1,055 LiveCodeBench problems according to whether all candidates in their five-candidate pools are incorrect, some but not all are correct, or all are correct.
On all-incorrect pools, where fixed-pool selection cannot succeed, VRR recovers 18 of 117 problems (15.38\%) with Gemma and 7 of 215 (3.26\%) with Qwen; each recovery therefore comes from a generated alternative selected after screening.

The initial-pool oracle covers 938 problems for Gemma and 840 for Qwen, corresponding to accuracies of 88.91\% and 79.62\%, respectively. VRR produces 938 and 839 correct final outputs, corresponding to accuracies of 88.91\% and 79.53\%, respectively.
Thus, Gemma's 18 recoveries are offset by 18 errors on covered problems, while Qwen's seven recoveries fall one short of eight covered-problem errors.
On mixed pools, VRR is correct on four fewer problems than \lav{} with Gemma and four more with Qwen; these complete-system differences do not represent observed VRR transitions.
The single Qwen error on an all-correct pool is, however, a genuine correct-to-incorrect regression.
These results demonstrate recovery beyond the initial pool.

\subsection{Operational Runtime}
\label{sec:exp-runtime}

In Table~\ref{tab:runtime}, we report recorded runtime to characterize operational cost rather than to provide a controlled comparison of algorithmic efficiency.
VRR has lower recorded method-specific runtime than \lav{} in all eight model--benchmark combinations. 
Because batching, caching, and parallelization differ across runs, these measurements do not constitute a controlled comparison of latency or compute efficiency.

\begin{table}[tbp]
  \centering
  \small
  \caption{Mean recorded runtime (seconds per problem); \lav{} and VRR exclude shared initial generation.}
  \label{tab:runtime}
  \begin{tabular}{llrrr}
    \toprule
    Model & Dataset & Initial generation & \lav{} & \cellcolor{vrrcolumn}VRR \\
    \midrule
    Gemma 4 31B IT & LiveCodeBench & 39.52 & 156.31 & \cellcolor{vrrcolumn}108.91 \\
    Gemma 4 31B IT & MMLU-Pro & 38.19 & 47.20 & \cellcolor{vrrcolumn}42.21 \\
    Gemma 4 31B IT & GPQA Diamond & 57.08 & 70.79 & \cellcolor{vrrcolumn}51.05 \\
    Gemma 4 31B IT & AIME & 101.85 & 104.12 & \cellcolor{vrrcolumn}86.90 \\
    \midrule
    Qwen3-32B & LiveCodeBench & 36.21 & 169.31 & \cellcolor{vrrcolumn}69.08 \\
    Qwen3-32B & MMLU-Pro & 37.35 & 58.42 & \cellcolor{vrrcolumn}31.89 \\
    Qwen3-32B & GPQA Diamond & 63.14 & 74.38 & \cellcolor{vrrcolumn}44.13 \\
    Qwen3-32B & AIME & 102.38 & 87.05 & \cellcolor{vrrcolumn}75.92 \\
    \bottomrule
  \end{tabular}
\end{table}

\section{Related Work}

\paragraph{Test-time scaling and candidate search.}
Test-time methods improve coverage by sampling reasoning paths or solutions \citep{wang2023selfconsistency,brown2024largelanguagemonkeys}.
Search-based methods instead alternate generation and heuristic evaluation \citep{yao2023treeofthoughts}.
Adaptive policies can outperform uniform best-of-$N$ sampling because each problem benefits from a different compute allocation \citep{snell2025scaling}.
VRR complements these methods by using structured verification results to trigger additional generation.

\paragraph{Verifiers and LLM judges.}
Outcome verifiers score final answers, while process verifiers score intermediate steps \citep{cobbe2021training,lightman2023verify}.
Other methods use language models for backward verification or train generative verifiers to produce judgments with rationales \citep{weng2023selfverification,zhang2025generative}.
LLM-as-a-Verifier is the closest method to our selection setting.
It ranks candidates with fine-grained scores, repeated evaluations, and decomposed criteria \citep{kwok2026llmverifier}.
Whereas LLM-as-a-Verifier ranks a fixed candidate pool, VRR reuses its fine-grained judgments to generate additional candidates before reselection.
It also uses an explicit \unk{} state for unresolved claims and compares the leading candidates in both presentation orders to reduce position bias \citep{wang2024unfairevaluators}.

\paragraph{Feedback-guided refinement and repair.}
Self-Refine and Reflexion use natural-language feedback to improve model outputs iteratively \citep{madaan2023selfrefine,shinn2023reflexion}.
CodeRL and Self-Debugging use execution feedback to guide code regeneration \citep{le2022coderl,chen2023selfdebug}.
These methods demonstrate that feedback can improve later attempts.
Our focus is the complementary question raised by verifier-based inference: how to reuse evidence originally produced for candidate ranking as a signal for candidate improvement.
VRR retains the selected candidate, generates alternatives with complementary roles, and reselects under the original criteria rather than treating a revision as an automatic replacement.

\section{Conclusion}

We introduced LLM-as-an-Improver and proposed VRR, which reuses verification evidence to generate additional candidates before final reselection.
Across two models and eight benchmarks, VRR outperforms fixed-pool \lav{} selection in nine of 16 settings, ties it in four, and underperforms it in three.
More importantly, when all five initial LiveCodeBench candidates are incorrect, VRR recovers correct solutions in 15.38\% of cases with Gemma and 3.26\% with Qwen, demonstrating that verification evidence can produce correct candidates beyond a fixed initial pool.

\bibliographystyle{iclr2027_conference}
\bibliography{iclr2027_conference}

\appendix
\section{Detailed Experimental Configuration}
\label{app:experimental-details}

\subsection{Models and Initial Candidates}

We run Gemma 4 31B IT \citep{gemmateam2026gemma4} and Qwen3-32B \citep{qwenteam2025qwen3} with vLLM in bfloat16 and disable thinking mode.
We sample initial candidates with temperature $1.0$, top-$p$ $0.95$, and a maximum output length of 4,096 tokens.
We generate five candidates per code problem and three per reasoning problem.
All methods share the same initial pool within each model.

\subsection{Benchmarks}

We use LiveCodeBench \citep{jain2024livecodebench} and two EvalPlus code benchmarks: HumanEval+ and the extension of Mostly Basic Python Problems (MBPP) \citep{liu2023evalplus}.
We use MMLU-Pro \citep{wang2024mmlupro}, GPQA Diamond \citep{rein2023gpqa}, the 2025 American Invitational Mathematics Examination (AIME) \citep{zhang2025aime25}, the 2023 American Mathematics Competitions (AMC) \citep{mathai2025amc23}, and MATH-500 \citep{lightman2023verify} for reasoning.

\subsection{VRR Configuration}

We set the initial verification budget to $B_{\mathrm{init}}=64$ and the reselection budget to $B_{\mathrm{reselect}}=96$.
The initial selector retains the top two candidates.
We set $\epsilon=10^{-6}$ and use soft criteria only when the hard-score difference is at most $\tau_{\mathrm{soft}}=0.01$.
We set $\tau_{\mathrm{sat}}=\tau_{\mathrm{viol}}=0.8$, $\tau_{\mathrm{score}}=\tau_{\mathrm{unk}}=0.5$, and $K_{\mathrm{unk}}=3$.
When triggered, the repair stage generates three candidates: one from the selected candidate, one from the runner-up, and one from a new approach.
We use temperature $0.7$, top-$p$ $0.95$, and a maximum output length of 8,192 tokens for repair.
We fix the reselection seed to 0.

\subsection{\lav{} Configuration}

\lav{} assigns continuous scores to candidate pairs and ranks them with a Probabilistic Pivot Tournament \citep{kwok2026llmverifier}.
For an evaluated pair $(c_i,c_\ell)$ and criterion $o_j$, the verifier returns a distribution over the ordered score tokens $V_{\mathrm{score}}=\{v_1,\ldots,v_G\}$.
Let $\widetilde p_{i\mid\ell,j}^{(t)}(v_g)$ denote the distribution extracted for $c_i$ in repetition $t$; its conditioning includes the task, both candidates, their presentation order, and the criterion.
The corresponding expected score is
\begin{equation}
  s_{i\mid\ell,j}^{(t)}
  =\sum_{g=1}^{G}
  \widetilde p_{i\mid\ell,j}^{(t)}(v_g)\phi(v_g),
  \label{eq:lav-expected-score}
\end{equation}
where $\phi(v_g)$ maps each token to its ordered numerical value.
Given $J$ evaluation criteria and $K$ repeated evaluations, \lav{} aggregates the pair-conditioned expectations as
\begin{equation}
  R_{i\mid\ell}
  =\frac{1}{JK}
  \sum_{j=1}^{J}\sum_{t=1}^{K}s_{i\mid\ell,j}^{(t)}.
  \label{eq:lav-aggregate-score}
\end{equation}
Here, $J$ is the number of evaluation criteria, $K$ is the number of repeated evaluations per pair and criterion, and $G$ is the number of ordered score tokens.
We normalize the averaged scores to $[0,1]$ before computing pairwise preferences.
This notation follows \citet{kwok2026llmverifier}.
Thus, $K=4$ means that the verifier judges each pair four times under each criterion, averages those scores, and then averages across the $J$ criteria.
Repeated evaluation is an aggregation choice; any variance reduction depends on the dependence among calls.
It does not change the number of candidates, pivots, or repair iterations.

The Bradley--Terry model converts the averaged scores into pairwise preference probabilities.
The tournament first evaluates candidates along a random ring.
This ring places each candidate once in each presentation position.
The tournament then selects the top $k$ candidates as pivots.
It compares every non-pivot with every pivot and also compares all pivot pairs.
For $N$ candidates, the procedure uses
\begin{equation}
  N+k(N-k)+\binom{k}{2}
  \label{eq:lav-ppt-comparisons}
\end{equation}
pairwise comparisons, which scales as $\mathcal{O}(Nk)$ for $k\ll N$.
This expression counts candidate pairs, not model calls or tokens across criteria and repetitions.
The method sums each candidate's preference mass and divides by its number of comparisons.
It returns the candidate with the highest normalized mass.

We use three criteria, $k=2$ pivots, and $K=4$ evaluations per candidate pair and criterion.

\subsection{Information Available to Each Method}

For code tasks, we provide only the problem statement, compilation results, and public tests.
For reasoning tasks, we provide only the problem statement, available answer choices, and required answer format.
We reveal correctness labels, reference answers, and private tests only after the method fixes its final answer.

\subsection{Evaluation}

For code tasks, we mark a candidate correct only if it passes every benchmark test.
For reasoning tasks, we normalize the final answer and compare it with the reference label.
For MATH-500 \citep{lightman2023verify}, \texttt{math\_verify} checks mathematical equivalence.

\section{Prompt Templates Used in the Experiments}
\label{app:prompts}

This section records the prompts used for generation, verification, repair, and
the \lav{} baseline. Text in angle brackets denotes a value inserted at run
time. Evidence catalogs and audit ledgers are serialized deterministically from
the corresponding task record. Line wrapping is changed for presentation, but
the instructions and output formats are preserved. Some experiment runners
used the earlier internal name ``FOVEA'' in a header; we replace that header by
``VRR'' below because the name itself carries no instruction or task information.
No prompt contains a reference answer, correctness label, private-test result,
or candidate reward.

\subsection{Initial Candidate Generation}

For HumanEval+ and MBPP, the user prompt is:

\begingroup\small
\begin{verbatim}
Solve the following <BENCHMARK> Python programming task.
Return one complete executable Python solution inside exactly one fenced
```python block. Preserve the requested function name and interface.
Do not include tests or explanations outside the code block.

<PROBLEM>
\end{verbatim}
\endgroup

For LiveCodeBench, we use the benchmark question followed by either its starter
code or a standard-input/standard-output instruction, and end with:

\begingroup\small
\begin{verbatim}
### Answer: (use the provided format with backticks)
\end{verbatim}
\endgroup

For GPQA Diamond, the system prompt is ``You are an expert graduate-level
science reasoner.'' For AIME, AMC, and MATH-500, it is ``You are an expert
competition mathematician.'' The corresponding user template is:

\begingroup\small
\begin{verbatim}
<QUESTION OR PROBLEM, INCLUDING OPTIONS WHEN APPLICABLE>

Solve independently and show a complete derivation.
<FINAL-ANSWER FORMAT INSTRUCTION>
\end{verbatim}
\endgroup

The final-answer instruction requires a separate
\texttt{FINAL\_ANSWER: X} line for multiple-choice tasks and a separate
\texttt{FINAL\_ANSWER: \textbackslash boxed\{answer\}} line for free-form
mathematics tasks.

\subsection{Candidate-Blind Criterion Generation}

Task-specific evaluation items are generated before any candidate is shown.
Human-authored dataset-level items are supplied in
\texttt{<FIXED OBLIGATIONS>} and remain part of the final contract.

\begingroup\small
\begin{verbatim}
You compile a candidate-blind success contract. You have NOT been
given any candidate answer or trajectory. Derive only atomic conditions
entailed by the task plus universal execution invariants. Fatal
requirements are hard; preferences are soft. Return JSON only:
{"obligations":[{"id":"snake_case","kind":"hard|soft",
"predicate":"one independently auditable condition",
"support":["required evidence types"],
"counterexample":["concrete falsifiers"],
"dependencies":[],"prior":0.5}]}
Do not reward effort, narration, or style. Include post-final-change
verification when the task changes executable state. The following
human-authored fixed conditions are already part of the final AND
contract and will be audited in parallel. You may overlap with them
when a task-specific clause is useful:
<FIXED OBLIGATIONS>
Return at most <MAX> concise task-specific obligations.
Compiler seed: <SEED>

TASK:
<PROBLEM>
\end{verbatim}
\endgroup

The fixed code criteria ask about specification and interface adherence,
functional correctness including boundary cases, and visible error signals such
as syntax errors, runtime failures, failed public checks, or infeasible
complexity. The fixed reasoning criteria ask about instruction and answer-format
adherence, correctness of material factual, mathematical, and logical steps, and
whether the reasoning entails the final answer.

\subsection{Batched Initial Verification}

The initial pass audits all items for one candidate in one call. The prompt
version used by the implementation is
\texttt{batched-candidate-svu-v3}.

\begingroup\small
\begin{verbatim}
You are a skeptical falsification auditor. Audit EVERY atomic
obligation for ONE candidate in this single response, using only that
obligation's allowed evidence records. Agent claims are not proof.
SAT needs concrete cited support. VIOL needs a concrete cited
counterexample. Missing, ambiguous, pre-change, invalidly cited, or
unavailable evidence is UNKNOWN. Never compare against another
candidate and never infer an unrecorded event.

TASK:
<PROBLEM>

AUDIT MODE: <MODE>

EVIDENCE CATALOG (id | type | exact raw span | time):
<EVIDENCE CATALOG>

<ONE BLOCK PER OBLIGATION, CONTAINING ITS ID, KIND, PREDICATE,
 EXPECTED SUPPORT, FALSIFIERS, AND ALLOWED EVIDENCE IDS>

Return exactly one block for every index, in ascending order. Cite only
ids listed for that obligation. Each verdict letter must be generated
in the response so its generated-position logits define the S/V/U
distribution.
<audit index="0"><reason>brief reason</reason>
<evidence>E000001</evidence><verdict>S</verdict></audit>
Use S=SAT, V=VIOL, U=UNKNOWN. Do not omit, duplicate, or add indexes.
\end{verbatim}
\endgroup

\subsection{Adaptive Atomic Verification}

When the acquisition rule requests another judgment, VRR uses the following
single-item prompt (implementation version \texttt{atomic-svu-v4}):

\begingroup\small
\begin{verbatim}
You are a skeptical falsification auditor. Decide one atomic success
obligation using ONLY the cited evidence records below. Agent claims
are not proof. SAT requires concrete supporting evidence. VIOL
requires a concrete counterexample. Missing, ambiguous, pre-change,
or uncited evidence is UNKNOWN. Never infer an event that is not present.

TASK:
<PROBLEM>

ATOMIC OBLIGATION [<ID>, <KIND>]:
<PREDICATE>

Expected support: <SUPPORT>
Concrete falsifiers: <FALSIFIERS>
Audit mode: <MODE>

EVIDENCE RECORDS (id | type | exact raw span | time):
<EVIDENCE RECORDS>

Return exactly these tags. Cite only listed ids. Use the one-letter
encoding S=SAT, V=VIOL, U=UNKNOWN so logits are a three-way distribution.
<reason>brief evidence-grounded reason</reason>
<evidence>E000001,E000002</evidence>
<verdict>S|V|U</verdict>
\end{verbatim}
\endgroup

The implementation obtains $\boldsymbol q_{ij}^{(t)}$ from the generated-position
log probabilities of the final verdict token. Unrecognized and omitted top-$k$
mass is assigned to \unk{}, and the three components are normalized to sum to
one. A \sat{} or \viol{} verdict without valid retrieved evidence is replaced
by \unk{}.

\subsection{Pairwise Reselection Prompt}

Finalists are compared in both presentation orders with prompt version
\texttt{paired-evidence-abt-v3}:

\begingroup\small
\begin{verbatim}
You are a skeptical final comparison auditor. Compare two candidates
for one atomic success obligation using ONLY the evidence records
below. A and B are anonymous positions, not quality hints. Agent
claims are not proof. Missing, ambiguous, pre-change, or uncited
evidence means T (tie/unknown). Never infer an event that is not present.

TASK:
<PROBLEM>

ATOMIC OBLIGATION [<ID>, <KIND>]:
<PREDICATE>

Expected support: <SUPPORT>
Concrete falsifiers: <FALSIFIERS>
Orientation: <ORIENTATION>

CANDIDATE-ONLY EDIT DELTA:
<DELTA>

OBLIGATION/DELTA ALIGNMENT:
<ALIGNMENT>

CANDIDATE A EVIDENCE:
<A-PREFIXED EVIDENCE>

CANDIDATE B EVIDENCE:
<B-PREFIXED EVIDENCE>

Return exactly these tags. Cite only listed candidate-prefixed ids and
cite at least one A record and one B record for an A/B preference.
A means candidate A has stronger evidence, B means candidate B, and
T means tie/unknown.
<reason>brief evidence-grounded comparison</reason>
<evidence>A:E000001,B:E000001</evidence>
<verdict>A|B|T</verdict>
\end{verbatim}
\endgroup

\subsection{Repair Generation}

Each Repair request receives one source candidate, except for a from-scratch
variant when no parent is supplied, together with a compact ledger containing
the criteria, initial ranking, scores, up to four strengths, up to four
violation hypotheses, and up to two unresolved items per candidate. Code tasks
use the following common wrapper:

\begingroup\small
\begin{verbatim}
You are generating one additional candidate for a label-blind VRR
portfolio-repair experiment. Private tests, candidate rewards, and
answer labels are unavailable. Return exactly one complete Python
solution inside a fenced ```python block and no other fenced block.

Role: <ROLE>
<ROLE-SPECIFIC INSTRUCTION>

Problem specification:
<PROBLEM>

<PARENT CANDIDATE OR "NO PARENT CANDIDATE IS SUPPLIED">

Public-evidence audit ledger:
<COMPACT LEDGER>

Requirements:
- Preserve the required interface and input/output contract.
- Handle boundary cases and the full stated constraints.
- Do not rely on private tests or candidate identities.
- Treat semantic audit claims as hypotheses, not ground truth.
- Before coding, privately establish the invariant and complexity bound.
- Check minimum, maximum, degenerate, and off-by-one cases.
- The result must compile and pass all designated public cases.
\end{verbatim}
\endgroup

The three role-specific instructions are:

\begingroup\small
\begin{verbatim}
SELECTED-CANDIDATE REPAIR:
Repair the parent while preserving every sound part. Privately derive
at least one concrete edge case for every applicable audit hypothesis,
then correct the root cause rather than its symptom.

RUNNER-UP REPAIR:
Repair the alternative parent after privately stress-testing its
invariants, asymptotic complexity, and boundary cases. Preserve sound
parts, but switch algorithms if the proof does not hold.

NEW-APPROACH SOLUTION:
Use the parent only as a bug-finding target. Privately derive a minimal
valid counterexample or boundary family on which it may fail, then
reimplement the solution from a correctness argument. Prefer an
algorithmically independent formulation and do not copy the parent's
control flow mechanically.
\end{verbatim}
\endgroup

For reasoning tasks, the common output portion is instead:

\begingroup\small
\begin{verbatim}
You are generating one additional candidate in a label-blind VRR
portfolio-repair experiment. The reference answer, correctness labels,
and benchmark evaluator are unavailable. Produce a complete,
self-contained solution.

Role: <ROLE>
<ROLE-SPECIFIC INSTRUCTION>

Problem:
<PROBLEM>

Parent candidate:
<PARENT>

Audit ledger (semantic audit claims are hypotheses, not ground truth):
<COMPACT LEDGER>

Requirements:
- Check the full problem and all answer choices or domain constraints.
- Do not vote among candidates or infer a hidden reference answer.
- Give enough reasoning that the final conclusion can be audited.
- <DATASET-SPECIFIC FINAL-ANSWER FORMAT>
\end{verbatim}
\endgroup

The role instructions tell the model, respectively, to preserve sound reasoning
while correcting ledger-indicated root causes, to independently check every
pivotal fact and inference in the runner-up, or to use the parent only as a
bug-finding target and solve again from first principles.

\subsection{LAV Pairwise Scoring Prompt}

The \lav{} runs use the frozen legacy prompt. The complete task and both
trajectories are inserted before the pre-specified criterion:

\begingroup\small
\begin{verbatim}
You are an expert evaluator of AI coding agents. You will see a task
description and two agent trajectories. Your job is to evaluate them
on ONE specific criterion: <CRITERION NAME>.

<GROUND-TRUTH AVAILABILITY NOTE>

Task:
<PROBLEM>

Trajectory A:
<CANDIDATE A>

Trajectory B:
<CANDIDATE B>

Evaluation Guideline -- <CRITERION NAME>:
<CRITERION DESCRIPTION>

Score each trajectory ONLY on this specific criterion. Ignore other
aspects that are not relevant to the criterion.

Rating Scale:
A = clearly and completely succeeded with verified output (best)
B-D = succeeded with only minor issues
E-G = above average, mostly correct with some issues
H-J = uncertain, leans toward success
K-M = uncertain, leans toward failure
N-P = below average, significant issues remain
Q-S = failed with some partial progress
T = clearly and completely failed (worst)

Then output your final scores:
<score_A>LETTER_A_TO_T</score_A>
<score_B>LETTER_A_TO_T</score_B>

Begin your analysis now.
\end{verbatim}
\endgroup

The expected score is computed from the token-level distribution over A--T,
not from only the emitted letter. As described above, we use three criteria,
two pivots, and four repetitions per pair and criterion.

\section{A Concrete All-Incorrect Recovery Example}
\label{app:all-incorrect-example}

We provide a concrete LiveCodeBench example in which VRR produces a correct solution despite all candidates in the initial pool being incorrect.
The task asks for the largest value obtainable from an array by repeatedly merging two adjacent values when the left value is no larger than the right value.
For example, the optimum for $[2,3,7,9,3]$ is 21: merging 7 into 9 first yields $[2,3,16,3]$, after which 3 and then 2 can be absorbed into 16.

All five initial candidates were incorrect under the benchmark tests.
Despite being sampled independently, they implemented essentially the same left-to-right greedy rule:
\begin{verbatim}
running_sum = nums[0]
answer = nums[0]
for x in nums[1:]:
    if running_sum <= x:
        running_sum += x
    else:
        running_sum = x
    answer = max(answer, running_sum)
\end{verbatim}
On the example above, this rule prematurely forms $2+3+7=12$, which cannot be merged into 9, and therefore returns 12 rather than 21.
Because \lav{} can only rank the five fixed candidates, it cannot recover on this problem.

The initial verification stage found that every candidate produced an incorrect answer on one of the two official public cases.
The resulting Failure Ledger preserved two distinct facts: the required class and method declarations were present, but functional correctness was violated because the left-to-right accumulation did not realize the optimal merge order.
Since every candidate fell below the repair threshold, VRR triggered candidate expansion and used the highest-ranked initial candidate as one repair parent.

During repair, the model explicitly reconsidered the order of operations: merging 5 into 7 produces 12 and blocks the subsequent merge into 9, whereas merging 7 into 9 first produces 16 and then permits the values to its left to be absorbed.
It therefore changed the traversal direction and generated the following core algorithm:
\begin{verbatim}
current = nums[-1]
answer = current
for i in range(len(nums) - 2, -1, -1):
    if nums[i] <= current:
        current += nums[i]
    else:
        current = nums[i]
    answer = max(answer, current)
return answer
\end{verbatim}
This right-to-left invariant represents the value already constructed to the right of position $i$; whenever \texttt{nums[i]} is no larger, it can be absorbed into that value.

The repaired candidate passed compilation and both public cases, was admitted to the expanded pool, and was selected during final reselection.
The private tests, evaluated only after selection, confirmed that the repaired candidate was correct.
\end{document}